\documentclass[letterpaper]{article} 
\usepackage{aaai25}  
\usepackage{times}  
\usepackage{helvet}  
\usepackage{courier}  
\usepackage[hyphens]{url}  
\usepackage{graphicx} 
\usepackage{natbib}  
\usepackage{caption} 
\usepackage{algorithm}
\usepackage{algorithmic}
\usepackage{subcaption}
\usepackage{amsmath} 
\usepackage{amssymb} 
\usepackage{amstext} 
\usepackage{multirow}
\usepackage{booktabs} 
\usepackage{epstopdf}
\newcommand{\resizeboxsize}{0.90}
\newcommand{\name}{LLM4NG}
\usepackage{enumitem}
\usepackage{newfloat}
\usepackage{listings}
\DeclareCaptionStyle{ruled}{labelfont=normalfont,labelsep=colon,strut=off} 
\floatstyle{ruled}
\newfloat{listing}{tb}{lst}{}
\floatname{listing}{Listing}
\title{Leveraging Large Language Models for Node Generation in Few-Shot Learning on Text-Attributed Graphs}
\author{
    Jianxiang Yu,
    Yuxiang Ren\thanks{Corresponding authors.},
    Chenghua Gong,
    Jiaqi Tan,
    Xiang Li\footnotemark[1],
    Xuecang Zhang
}
\affiliations{
    \textsuperscript{\rm 1}East China Normal University\\
    \textsuperscript{\rm 2}Advance Computing and Storage Lab, Huawei Technologies \\

    \{jianxiangyu, chenghuagong, jiaqitan\}@stu.ecnu.edu.cn,
    \{renyuxiang1,zhangxuecang\}@huawei.com,
    xiangli@dase.ecnu.edu.cn

}

\usepackage{bibentry}

\begin{document}

\maketitle

\begin{abstract}
Text-attributed graphs have recently garnered significant attention due to their wide range of applications in web domains. Existing methodologies employ word embedding models for acquiring text representations as node features, which are subsequently fed into Graph Neural Networks (GNNs) for training. Recently, the advent of Large Language Models (LLMs)  has introduced their powerful capabilities in information retrieval and text generation, which can greatly enhance the text attributes of graph data. Furthermore, the acquisition and labeling of extensive datasets are both costly and time-consuming endeavors. Consequently, few-shot learning has emerged as a crucial problem in the context of graph learning tasks. In order to tackle this challenge, we propose a lightweight paradigm called \name, which adopts a plug-and-play approach to establish supervision signals by leveraging Large La
nguage Models (LLMs) for node generation. Specifically, we utilize LLMs to extract semantic information from the labels and generate samples that belong to these categories as exemplars. Subsequently, we employ an edge predictor to capture the structural information inherent in the raw dataset and integrate the newly generated samples into the original graph. This approach harnesses LLMs for enhancing class-level information and seamlessly introduces labeled nodes and edges without modifying the raw dataset, thereby facilitating the node classification task in few-shot scenarios. Extensive experiments demonstrate the outstanding performance of our proposed paradigm, particularly in low-shot scenarios. For instance, in the 1-shot setting of the ogbn-arxiv dataset, \name~achieves a 76\% improvement over the baseline model.
\end{abstract}

%

\section{Introduction}
Text-Attributed Graphs (TAGs) are prevalent in a variety of real-world scenarios, such as product networks, social networks, and citation networks~\cite{yao2019graph,nguyen2020fang}.
In TAGs,
nodes represent entities with textual information and edges capture relationships between entities.
For example,
amazon product network is one of the web applications
and in its review datasets
~\cite{ni-etal-2019-justifying},
each product can be represented as a node,
featured with various text attributes, such as descriptions and brands.
By constructing edges between nodes based on user purchase history, 
TAGs effectively capture the relationships among different products.
TAGs are a powerful representation of data that combines textual information with graph structures. 
The utilization of TAGs empowers us to unlock new discoveries across various domains, including recommendation systems~\cite{zhu2021textgnn} and fake news detection
~\cite{benamira2019semi}.

Graph Neural Networks (GNNs) currently utilize graph structure and node features to learn the representation of nodes through message propagation strategies.
Many GNNs adopt
a naive way to {encode} the textual information of nodes in TAGs as non-contextualized shallow embeddings~\cite{miaschi-dellorletta-2020-contextual} e.g., Bag-of-Words~\cite{bow} and TF-IDF~\cite{tfidf}~embeddings.
However, 
the resulting embeddings are unable to capture polysemy and the semantic relationships between words.
Recent advancements in natural language processing (NLP) have introduced contextualized word embeddings such as BERT~\cite{bert} and Deberta~\cite{he2020deberta}. 
In particular, 
Sentence-BERT~\cite{sentence-bert} performs better in sentence-level text representations.
These language models (LMs) capture the contextual information of words and sentences, leading to more powerful representations of text.
There have been some GNNs
that utilize both LMs and GNNs for training, combining text features and graph topology to obtain {effective} node representations~\cite{glem,graphformers}.

The emergence of Large Language Models (LLMs)
like GPT~\cite{gpt3},
Llama~\cite{touvron2023llama}
and ChatGLM~\cite{du2021glm},
has made a significant impact due to their powerful generative capabilities.
These models typically have a large number of parameters.
They can acquire rich linguistic knowledge and semantic representations
through training on large-scale corpora.
LLMs use prompt engineering to guide their generation and inference capabilities, mining the extensive knowledge they have learned. 
As a result, they exhibit exceptional performance across various natural language processing tasks, including machine translation, sentiment analysis, and contextual understanding.
Following their success,
incorporating LLMs into various applications has received wide attention~\cite{li2023empowering,fan2023recommender}.
In the field of GNNs,
\citet{kea} and \citet{tape} have made significant progress in exploring the potential of LLMs by enhancing node-level text in different ways.
However,
such methods can lead to much time consumption on large-scale graphs.
This is because
constantly invoking LLMs for each node poses efficiency challenges.
Furthermore,
in the few-shot learning scenarios,
the rich node features generated by LLMs could bring only marginal performance improvement due to the scarcity of labeled nodes.
Consequently, it is necessary to propose a method that ensures low consumption and is suitable for scenarios with extremely limited labeled data.

Intuitively,
as the training set becomes larger, it may contain more diverse information, which enables the models to perform better.
We expect to leverage LLMs to establish supervision signals and enhance the training dataset in TAGs,
rather than performing data augmentation directly on the nodes.
In the field of graph learning, different datasets correspond to various domains. 
For example,
for social networks, we need to know the relationship between user interests and user characteristics,
while for citation networks in computer science, we need to know expertise in software, operating systems, etc.
Therefore,
we need to utilize specific domain knowledge to capture the relationship between samples and labels.
LLMs can be regarded as comprehensive ``encyclopedias'' 
that encompass domain-specific knowledge across various fields.
We can exploit LLMs to provide robust support for models in text-attributed graph learning, helping to establish supervision signals and thus enhance the model performance.

Based on the above,
we propose a plug-and-play, lightweight paradigm named
\name~
that leverages \textbf{LLM}s \textbf{for} \textbf{N}ode \textbf{G}eneration.
Specifically,
the paradigm mines label semantics through LLMs,
generating diverse samples from different classes to perform class-level augmentation on the raw dataset, 
thereby obtaining supervision signals to assist GNNs in downstream few-shot node classification tasks.
We simply generate labeled samples from LLMs 
and insert them into the raw dataset to train them together on any GNNs.
In practice,
when dealing with node classification tasks in few-shot scenarios of a TAG,
using LLMs to generate samples presents two main challenges:
1) How to generate diverse and meaningful labeled samples?
2) How to effectively integrate the generated samples with the raw dataset?
Firstly, 
we utilize prompt engineering techniques to guide LLMs in mining semantic information from labels and generate samples of different classes.
The generated samples adhere to the textual format of nodes in the raw dataset.
In addition, 
we utilize different prompt statements to explore the diversity of generated samples and analyze how the quantity of generated samples and LLMs' stochasticity affect the quality of the generated samples.
Secondly, 
we train an edge predictor that uses the raw graph structure as supervision signals to establish connections between the generated nodes and nodes in the raw dataset. 
In this way,
we can seamlessly integrate the generated nodes into the topological structure of the raw dataset,
extending the diversity of the dataset and the size of the training set.
By employing these strategies, 
we can effectively address the challenges mentioned above.
Next, we can use any GNNs architecture, 
taking the merged node features and topological structure as input, 
to train and obtain classification results for the samples from the raw dataset. 
Using LLMs to generate nodes allows for a certain degree of fault tolerance because the focus is placed on the expertise embedded within the generated samples rather than their authenticity.
\name~enables us to harness the power of LLMs and enhance the overall performance of the model.
Finally, we summarize our contributions as:
\begin{itemize}[leftmargin=*, itemsep=0pt, parsep=0pt, topsep=0pt, partopsep=0pt]
\item We propose the use of LLMs for generating samples in graph domain to augment the raw dataset and enrich the training set.
To the best of our knowledge,
this is the first to use LLMs for node generation.
We also provide a new solution for training GNNs in zero-shot scenarios,
and it demonstrates superior performance.
\item Our proposed paradigm~\name~is plug-and-play, requiring only class-level invocation of the LLM and lightweight training on the structure,
which can be compatible with any GNNs.
Furthermore,
we integrate the labeled data generated by LLMs into the raw dataset.
\item Extensive experiments have shown that \name~exhibits remarkable performance.
While keeping the model and dataset unchanged, 
our paradigm significantly enhances the model's performance.
Notably,
in the 1-shot setting of the ogbn-arxiv dataset, \name~achieves a 76\% improvement over the baseline model.
\end{itemize}

\section{Related Work}
\begin{figure*}[t]
\centering
\includegraphics[width=0.95\linewidth]{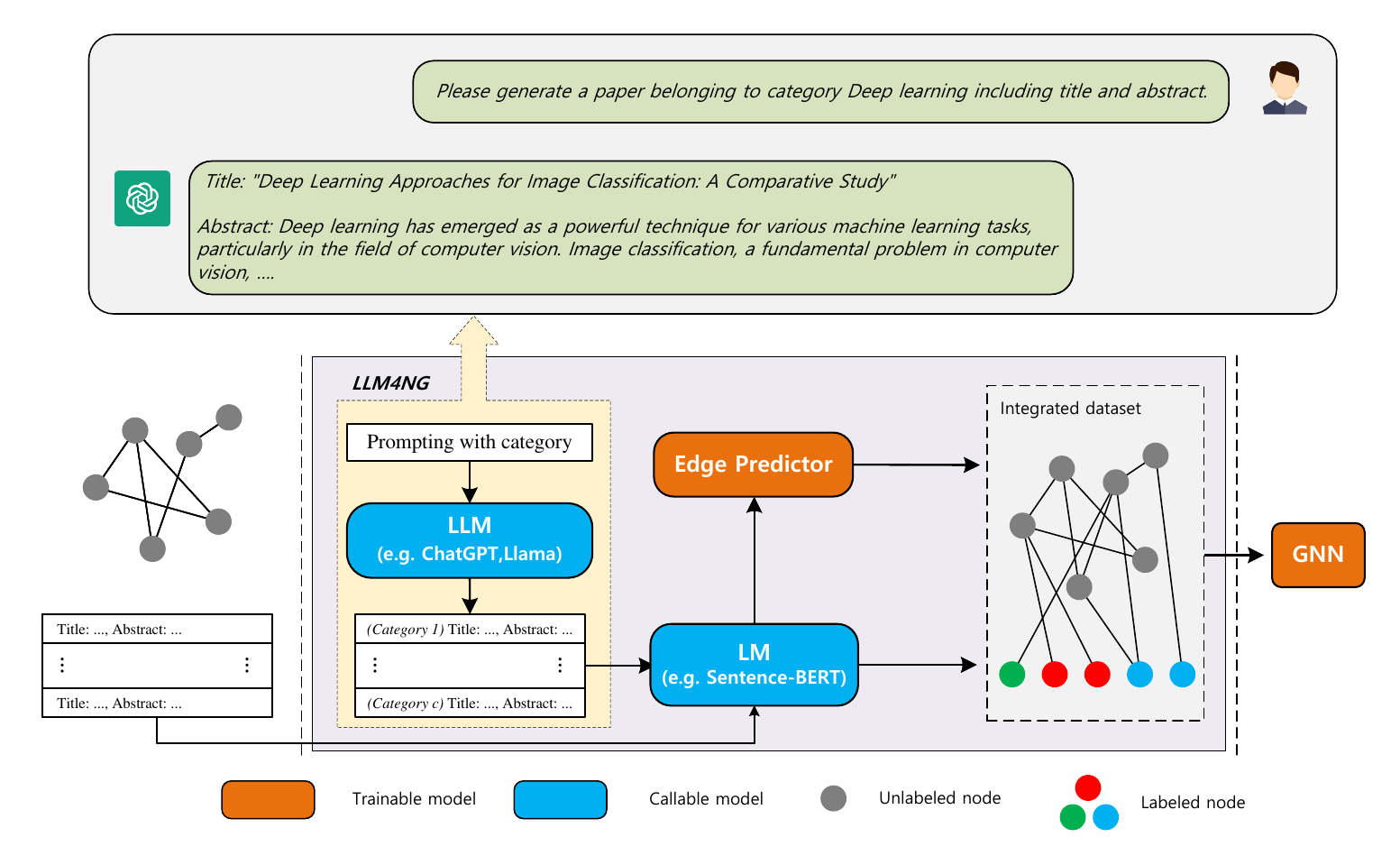}
\caption{The overall paradigm of LLM4NG.}
    \label{fig:model}
\end{figure*}

\paragraph{Language Models on TAGs.}
When dealing with text-attributed graph data, 
GNNs often adopt the approach of processing text as shallow non-contextual representations.
However, language models have a better understanding of contextual relationships and can express textual content more effectively.
Therefore, it becomes necessary to combine language models with graph neural networks to accomplish downstream tasks.
For example, 
GIANT~\cite{giant} extensively exploits the potential connections between graph structure and node attributes. 
It utilizes graph information to aid in feature extraction while fine-tuning language models. 
Graphformer~\cite{graphformers} nests graph aggregation within Transformer blocks for text encoding, 
enabling a global understanding of the semantic meaning of each node.
Further, 
GLEM~\cite{glem} trains graph neural networks and language models separately within a variational Expectation-Maximization framework, treating them as a process of knowledge distillation that mutually reinforces each other.
With the emergence of LLMs, 
we can leverage more extensive knowledge and information. 
TAPE~\cite{tape} utilizes LLMs to predict the ranked classification list of each node and outputs the corresponding explanations.
Additionally, KEA~\cite{kea} describes knowledge entities within the original text of each node to complement the textual content of the nodes.
Although these methods have achieved some success, 
they often come with high computational or invocation costs and may not effectively address the challenges posed by few-shot scenarios with insufficient supervision signals.

\paragraph{Few-shot learning on Graphs.}
In the case of real-world graphs, 
acquiring labels is a resource-intensive task, making it a formidable challenge in few-shot scenarios. 
Some methods adopt the paradigm of meta-learning to handle few-shot problems on graphs~\cite{wang2022task,lan2020node,liu2021relative,huang2020graph,liu2022few}.
In this paper, 
we focus on the semi-supervised setting, 
where each category has a few labeled samples.
Self-supervised learning methods on graphs learn node representations through a pre-training task and fine-tuning for classification results using a few labeled nodes~\cite{wan2021contrastive,grace,dgi,hafidi2020graphcl,wang2017mgae}.
For example,
GRACE~\cite{grace} and GraphCL~\cite{hafidi2020graphcl} 
construct different views through data augmentation on graphs to bring positive samples closer and push negative samples further apart for learning node representations. 
Additionally, DGI~\cite{dgi} stands as the pioneering method to introduce contrast between node-level embeddings and graph-level embeddings. This innovation empowers graph encoders to capture both local and global semantic information.
MVGRL~\cite{mvgrl} leverages first-order neighbor subgraphs and graph diffusion to generate contrastive views.
However, when labeled data is extremely scarce leading to insufficient supervision signals, 
the model's performance suffers.
Several semi-supervised methods have been proposed to enhance model performance using limited labeled information. 
M3S~\cite{m3s} combines self-training and deep clustering to expand the training set by predicting high-confidence pseudo-labels.
CGPN~\cite{cgpn} establishes a variational inference framework for Graph Poisson Network~(GPN) and GNN models,
and integrates contrastive learning to propagate limited label information over the entire graph.
Meta-PN~\cite{metapn} generates high-quality pseudo-labels through meta-learning strategies to effectively enhance sparse label data.
These methods leverage the information of the raw dataset
to establish high-confidence pseudo-labels.
The advent of LLMs provides us with a novel approach to address few-shot problem.
We can harness these LLMs to unearth domain knowledge and generate extra supervision signals.

\section{Preliminary}

\paragraph{Text-Attributed Graphs.}
A Text-Attributed Graph (TAG) is defined as $\mathcal{G}_S=(\mathcal{V},\mathcal{E}, \mathcal{S}^{\mathcal{V}})$,
where $\mathcal{V}$ represents a set of nodes and
$\mathcal{E} \subseteq \mathcal{V} \times \mathcal{V}$ represents a set of edges.
Each node $v_i \in \mathcal{V}$ is associated with a sequential text $s_i \in \mathcal{S}^{\mathcal{V}}$
and
the corresponding label $y_i$.
Each label has a real label text
$c$ 
(e.g., `Machine Learning' or `Databases') from the set of all label texts $\mathcal{C}$.
The corresponding adjacency matrix of the graph $\mathcal{G}_S$ is denoted by $A$.
For simplicity, we set $A_{ij}$ = 1 if 
$e(v_i, v_j)
\in \mathcal{E}$ 
; 0, otherwise.


\paragraph{Few-shot Node Classification.}
In this paper, we focus on zero- and few-shot node classification, 
which is one of the most challenging and cutting-edge problems in the field of graph learning.
Unlike the few-shot problem in the meta-learning paradigm,
in our setting,
each class has $K$ labeled samples.
Given the labeled $K \times |\mathcal{C}|$ samples,
our goal is to predict unlabeled samples in the test set,  
referred to as $K$-shot classification.
Specifically, 
zero-shot classification means that there is no labeled sample.

\paragraph{Graph Neural Network Architecture}
GNNs use input graph structure as the computation graph, 
aggregating information from a node’s neighbors, then updating the representation of each node.
Formally, suppose $h^{(l)}_{v_i}$ is the representation of node $v_i$ at the $l$-th GNN layer, 
the updating procedure 
is:

\begin{equation}
h^{(l)}_{v_i} = 
\underset{ \forall v_j \in \mathcal{N}(v_i),  \forall e \in \mathcal{E}(v_i, v_j)}{\text{Aggregate}}
\left(
\{
\text{Propagate}(h^{(l-1)}_{v_i};h^{(l-1)}_{v_j},e)
\}
\right)
\end{equation}
where $\mathcal{N}(v_i)$ denotes the set of neighbors of node $v_i$ and $\mathcal{E}(v_i,v_j)$ denotes the set of edges connected from node $v_j$ to node $v_i$.
Aggregate$(\cdot)$ and Propagate$(\cdot)$ are the two functions in GNNs.
Propagate$(\cdot)$ represents the propagation of node information. 
It uses the target node’s representation $h^{(l-1)}_{v_i}$ and the edge $e$ between the two nodes as query, 
and propagates the information $h^{(l-1)}_{v_j}$ of the source node to the target node $v_i$. 
Aggregate$(\cdot)$ is a differentiable function (e.g., sum, mean, etc.) that aggregates the representations of a node’s neighbors.

\section{The Proposed Paradigm}

This section introduces details of our proposed paradigm~\name.
The general model diagram is
shown in Figure~\ref{fig:model}.
We guide LLM through prompt statements to generate some samples with labels and input these generated samples into  LM together with the text in the raw dataset to obtain node representations.
Subsequently, 
we perform supervised learning using the raw graph.
We utilize the existing edges as supervision signals and input node embeddings to train the edge predictor.
After training, we can utilize the embeddings to predict edges between nodes in the raw dataset and generated nodes.
The newly generated nodes, along with their structural information, can be integrated into the original dataset.
Finally, we can train the entire dataset using any GNNs and obtain results for node classification.

\subsection{Sample Generation from LLM}

When dealing with node classification tasks of a TAG in few-shot scenarios, 
we lack sufficient supervision signals to make accurate predictions. 
To address this, 
we can leverage the textual information within the set of label texts $\mathcal{C}$ to explore the semantics embedded in LLMs and generate $M$ relevant sample instances for each label.
We generate a total of $M \times |\mathcal{C}|$ samples,
$M \times |\mathcal{C}| \ll |\mathcal{V}|$,
which is less costly compared to node-level augmentation.
Specifically,
we denote $c$ as one of the label texts.
$\text{Prompt}(c)$ denotes 
the prompt statement containing the label text $c$.
We feed it into LLM and then get the generated text $s_g$:
\begin{equation}
    s_{g}=\text{LLM}(\text{Prompt}(c)),\quad
    c \in \mathcal{C}
\end{equation}

The text generated by LLM is based on the knowledge learned from the training corpora.
It contains both samples that actually exist in the real world and pseudo-samples that the model creates internally by permutating the knowledge.
Even though those pseudo-samples do not exist in reality,
we can still trust them because our goal is to capture domain knowledge associated with the label.
For instance, 
when performing interest classification of users in a social network, 
for the label ``sports'',
we can generate user profiles of individuals who are interested in basketball or soccer.
Although these users do not actually exist, 
we can still uncover domain-specific information and capture the information that basketball and soccer related to the ``sports'' category.

\subsection{Node representation initialization with LM}
After generating new samples, 
we treat each of them as a new node $g_i$ on the TAG.
Inspired by~\cite{kea},
deep sentence embeddings with GNNs have emerged as a powerful baseline compared to non-contextualized shallow embeddings. 
Moreover, sentence embedding models offer a lightweight approach to get representations without fine-tuning. 
Therefore, we adopt sentence embedding models to extract information from the text such as Sentence-BERT~\cite{sentence-bert} and e5-large~\cite{e5}.
\begin{equation}
h_{v_i} =\text{LM}(s_{v_i}),
\quad
h_{g_i}=\text{LM}(s_{g_i}),
\end{equation}
    where $s_{v_i}$ and $s_{g_i}$ denote the text of the raw dataset samples and the newly generated sample, respectively.
$h_{v_i}$ and $h_{g_i}$ denote the corresponding representations of these samples.
We denote $H_n=H_v \cup H_g$ as the representation of all nodes.

\subsection{Edge Predictor}

Next, we insert new nodes into the raw graph to establish connections between newly generated nodes $g_i$ and the raw nodes $v_j$.
Edges play a role in message passing in the paradigm of GNNs,
and new nodes with labels need to propagate supervision signals to other nodes without labels. 
We aim to ensure that unlabeled nodes receive as much relevant information from similar samples as possible to assist downstream classification tasks.

We consider this problem from two perspectives.
First, we perform a coarse screening of potential edges between new nodes and raw nodes.
To achieve this, we filter irrelevant edges by the cosine similarity of the representations in the latent space.
\begin{equation}
e(h_{g_i},h_{v_j}) \in \mathcal{E}_P,\quad
\text{if} \:
\text{sim}(h_{g_i},h_{v_j})=h_{g_i} h_{v_j}^T > \delta
\label{sim}
\end{equation}
where $\delta$ as a hyper-parameter represents the similarity threshold.
We add edges 
that have larger similarity weights than $\delta$ to 
the edge prediction set $\mathcal{E}_P$.

However, relying solely on node similarity to determine edge existence is too simple. 
It is also expected the newly added edges
share high similarity with raw edges in the graph.
To achieve this goal,
we use the edges in the raw graph as supervision signals and construct an edge predictor for the link prediction task. 
Specifically, 
we create a binary classification task where the training set is constructed from the raw graph, 
and the test set is the 
edge prediction set $\mathcal{E}_P$.
We take raw edges in the graph as positive samples while
randomly sampling an equal number of non-existent edges as negative samples:
\begin{equation}
y_e(h_{v_i}, h_{v_j}) = \left\{
\begin{aligned}
&\: 1 \quad \text{if } e(h_{v_i}, h_{v_j}) \in \mathcal{E} \\
&\: 0 \quad \text{if } e(h_{v_i}, h_{v_j}) \notin \mathcal{E} \\
\end{aligned}
\right.
\end{equation}
Then,
we concatenate the representations of two-end nodes in each edge and feed them into a multi-layer perceptron (MLP) to obtain the probability of edge existence $\hat{y}_e$.
\begin{equation}
\hat{y}_e(h_{v_i}, h_{v_j}) = \text{MLP}(h_{v_i} || h_{v_j})
\end{equation}
Finally, 
we treat $y_e$ as ground truth and use the Cross-Entropy loss as the objective function.
After training the edge predictor, 
we input the node pairs from the test set into the model to obtain the edge probabilities for the node pairs in the test set. 
We then select the top-$k$ edges with the highest probabilities and add them to the raw graph.
This integrates the newly added nodes with the raw graph.

\subsection{\name}
Finally, we derive the new adjacency matrix $A_n$ as well as the representations of nodes $H_n$,
which are further fed into any GNNs along with the set of labeled nodes $\mathcal{Y}_L$ (a total of $(K+M)*|\mathcal{C}|$ samples) to output the classification results $Y_n$:
\begin{equation}
Y_n = \text{GNN}(A_n, H_n, \mathcal{Y}_L).
\end{equation}
The proposed paradigm~\name~utilizes LLMs to mine semantic information of labels in few-shot scenarios, 
enabling sample generation and the establishment of supervision signals. Subsequently, 
we employ a simple model Edge Predictor to obtain additional structural information.
Our proposed approach
is lightweight, which allows us to 
facilitate node classification in few-shot scenarios by simply adding labeled nodes and edges without altering the raw dataset.

\section{Experiments}

\subsection{Datasets and Baselines}
To evaluate the performance of~\name, we employ three
citation network datasets:
Cora~\cite{cora}, Pubmed~\cite{pubmed}, and ogbn-arxiv~\cite{arxiv}.
Detailed descriptions and statistics on these datasets are provided in Supplement~S1.1.
We also compare~\name~with 9 other SOTA baselines,
which can be categorized into four groups:

\noindent$\bullet$
\textbf{[Classical base models]}: GCN~\cite{gcn} and GAT~\cite{gat} are common benchmark models in graph learning for node-level representation.

\noindent$\bullet$
\textbf{[Graph self-supervised models]}:
DGI~\cite{dgi}, MVGRL~\cite{mvgrl}, and GRACE~\cite{grace} are prominent benchmark approaches in graph neural networks for self-supervised learning tasks. 

\noindent$\bullet$
\textbf{[Traditional graph data augmentation methods]}:
Mixup~\cite{wang2021mixup} and
DropEdge~\cite{rong2019dropedge} are classic methods for augmenting graph data.

\noindent$\bullet$
\textbf{[Graph models in few-shot scenarios]}:
CGPN~\cite{cgpn} and Meta-PN~\cite{metapn} are proposed for few-shot scenarios.

When running GCN and GAT,
we execute the features built into PyG~\cite{pyg} and the sentence embeddings obtained from LM each once. 
Further, for more details on the dataset and baseline, 
see Appendix A.

\begin{table*}[t]
  \caption{Node classification performance in few-shot scenarios of Cora w.r.t. the classification accuracy.
  We highlight the best results in bold.
}
\centering
  \label{cora_classification}
  \resizebox{\resizeboxsize \linewidth}{!}{
  \begin{tabular}{c|cccccccc}
    \hline
    Methods &  0-shot 
    & 1-shot & 2-shot & 3-shot & 4-shot & 5-shot & 10-shot \\
    \hline
    GCN(Pyg) & - &
    60.32(4.12) &
64.86(5.83) &
68.30(4.81) &
71.94(1.85) &
75.74(1.81) &
78.16(2.57) &
\\
    GCN(MiniLM) & - &
64.26(6.22) &
69.28(5.56) &
74.02(4.70) &
75.64(2.93) &
78.30(1.15) &
80.24(1.58) 
\\
    GAT(Pyg) & - &
51.76(4.97) &
55.90(4.61) &
61.16(5.32) &
67.74(2.14) &
71.26(3.06) &
75.38(3.26) 
\\

GAT(MiniLM) & - &
60.02(6.24) &
67.58(4.18) &
72.14(5.46) &
74.24(2.37) &
76.90(1.61) &
79.16(1.20) 
\\


\hline
DGI &- &
62.06(7.40) &
71.26(6.02) &
72.96(2.88) &
76.88(3.50) &
78.42(2.11) &
79.92(1.25) 
\\
MVGRL & - &
61.94(10.64) &
67.48(4.55) &
67.68(1.83) &
74.66(1.96) &
76.98(3.27) &
79.16(1.97)
\\
GRACE & - &
67.00(9.17) &
68.78(3.98) &
73.20(5.87) &
75.36(2.33) &
77.68(2.34) &
80.12(1.03) 
\\
\hline
Mixup+GCN & - &
43.56(12.47) &
53.58(8.58) &
64.38(3.57) &
65.58(6.45) &
68.72(4.46) &
71.50(3.04) 
\\
DropEdge+GCN & - &
48.67(2.85) &
52.73(4.17) &
63.03(7.00) &
69.33(2.23) &
70.13(2.82) &
79.55(0.96) &
\\
\hline
Meta-PN & - &
59.20(10.99) &
68.92(4.74) &
75.30(1.80) &
76.50(2.63) &
77.10(3.37) &
80.12(1.15) 
\\
CGPN & - &
67.94(4.09) &
73.46(2.78) &
75.76(0.91) &
75.70(1.80) &
77.46(2.63) &
77.80(0.61)
\\

\hline
\name+GCN &
\textbf{74.46(1.60)} &
\textbf{76.48(0.90)} &
\textbf{76.08(2.75)} &
76.26(2.88) &
\textbf{76.94(2.07)} &
79.06(0.75) &
80.40(1.01) 
\\

\name+GAT &
74.14(1.22) &
75.90(1.86) &
75.44(2.38) &
\textbf{76.34(2.63)} &
76.40(1.51) &
78.14(1.25) &
79.60(1.16) 
\\

\name+GCN w/o A &
70.44(1.72) &
75.12(1.60) &
75.24(1.88) &
76.04(3.21) &
76.86(1.58) &
\textbf{79.28(1.20)} &
\textbf{80.94(1.55)} 
\\

\name+GAT w/o A  &
74.08(1.49) &
74.38(2.03) &
75.00(4.36) &
74.82(3.77) &
75.98(3.42) &
77.20(3.36) &
78.82(2.57) 
    \\
(improv.) & - & (+12.57\%) &
(+3.57\%) &
(+0.77\%) &
(+0.08\%) &
(+1.10\%) &
(+1.02\%)
\\
    \hline
  \end{tabular}
}
\end{table*}

\begin{table*}[t]
  \caption{Node classification performance in few-shot scenarios of Pubmed w.r.t. the classification accuracy.
  We highlight the best results in bold.}
  \centering
  \label{pubmed_classification}
   \resizebox{\resizeboxsize \linewidth}{!}{
  \begin{tabular}{c|ccccccc}
    \hline
    Methods &  0-shot 
    & 1-shot & 2-shot & 3-shot & 4-shot & 5-shot & 10-shot\\
    \hline
    GCN(Pyg) & - &
    60.52(3.15) &
    63.16(2.45) &
    63.22(4.59) &
    65.58(3.58) &
    68.64(3.15) &
    75.26(2.48) 
    \\

    GCN(MiniLM) & - &
    63.02(3.71) &
    64.74(4.00) &
    65.82(2.24) &
    68.40(4.16) &
    70.06(4.33) &
    77.72(1.56) 
    \\
    GAT(Pyg) & - &
    59.72(4.14) &
    63.48(3.15) &
    63.76(2.87) &
    65.88(3.80) &
    66.92(3.84) &
    74.64(3.55) 
    \\

     GAT(MiniLM) & - &
    61.82(4.21) &
    63.22(5.70) &
    65.38(2.30) &
    66.56(2.96) &
    69.22(1.47) &
    76.02(1.86) 
    \\

\hline
DGI & - &
64.88(7.84) &
64.70(9.13) &
69.88(3.39) &
70.98(3.72) &
73.76(3.96) &
77.96(0.95) 
\\
MVGRL &- &
61.78(9.22) &
64.7(10.64) &
65.42(8.10) &
69.36(1.08) &
69.58(3.08) &
75.16(4.41) 
\\
GRACE & - &
63.80(1.93) &
67.70(6.22) &
68.74(6.06) &	
69.60(5.67) &
71.46(7.00) &
76.86(3.18) 
\\
\hline
Mixup+GCN &- &
54.14(3.77) &
54.48(7.05) &
61.22(5.64) &
62.30(6.96) &
63.08(8.09) &
68.08(4.12) 
\\
DropEdge+GCN & - &
51.77(8.43) &
64.76(2.95) &
64.35(5.82) &
67.43(1.33) &
66.43(1.69) &
76.70(1.04) 
\\
\hline
Meta-PN & - &
57.52(3.85) &
59.56(6.16) &
66.60(7.24) &
69.52(9.38) &
69.66(6.55) &
74.28(4.84) 
\\

CGPN &- &
59.03(10.05) &
56.90(9.95) &
63.00(11.36) &
65.03(4.37) &
64.73(7.68) &
64.70(3.72)
\\

\hline
\name+GCN &
\textbf{75.36(2.43)} &
\textbf{75.06(2.56)} &
\textbf{76.82(4.30)} &
\textbf{76.70(3.46)} &
\textbf{76.74(2.55)} &
\textbf{78.50(2.56)} &
\textbf{80.54(0.78)} \\

\name+GAT &
74.66(1.04) &
75.98(2.82) &
76.50(1.85) &
76.06(2.67) &
76.00(3.15) &
78.04(2.22) &
78.66(1.07) 
\\

\name+GCN w/o A & 
73.72(2.88) &
74.82(2.45) &
75.56(3.17) &
75.72(3.36) &
76.06(2.68) &
77.66(2.85) &
79.64(1.33) 
\\

\name+GAT w/o A &
74.08(1.49) &
74.38(2.03) &
75.00(4.36) &
74.82(3.77) &
75.98(3.42) &
77.20(3.36) &
78.82(2.57) 
\\
(improv.) & - & (+15.70\%) &
(+13.47\%) &
(+9.76\%) &
(+8.11\%) &
(+6.42\%) &
(+3.31\%)
\\

    \hline
  \end{tabular}
}
\end{table*}

\begin{table*}[t]
\centering
  \caption{Node classification performance in few-shot scenarios of ogbn-arxiv w.r.t. the classification accuracy.
  We highlight the best results in bold.
  OOM denotes the out-of-the-memory error.}
  \label{arxiv_classification}
  \resizebox{\resizeboxsize \linewidth}{!}{
  \begin{tabular}{c|ccccccc}
    \hline
    Methods &  0-shot 
    & 1-shot & 2-shot & 3-shot & 4-shot & 5-shot & 10-shot \\
    \hline
    GCN(Pyg) & - & 
    23.38(5.26) & 
    32.98(9.12) & 
    39.77(6.36) & 
    46.23(3.62) &
    47.95(2.24) &
    52.27(1.86)  
    \\

    GCN(MiniLM) & - &
    31.42(3.89) &
    40.63(10.52) &
    45.62(7.02) &
    55.30(4.08) &
    53.44(2.01) &
    56.48(1.24) 
    \\

    GAT(Pyg) & - &
    22.83(4.48) &
    24.82(6.01) &
    29.70(3.03) &
    40.39(4.03) &
    40.17(3.47) &
    50.13(2.28) 
\\

    GAT(MiniLM) & - &
    32.19(5.17) &
41.43(8.28) &
45.03(5.66) &
53.92(2.35) &
51.94(1.80) &
54.39(1.22) 
\\

\hline

DGI & - &
25.75(5.97) &
32.93(6.30) &
36.47(7.78) &
40.71(2.42) &
42.32(3.11) &
45.79(1.27) 
\\
MVGRL & - &
16.24(1.71) &
17.82(2.26) &
21.58(4.11) &
23.42(1.79) &
24.50(1.45) &
31.26(1.17) 
\\
GRACE & - &
29.91(1.73) &
35.72(2.35) &
38.24(3.62) &
38.68(1.11) &
37.72(2.11) &
43.51(0.91) 
\\
\hline
Mixup+GCN & - &
20.52(5.95) &
27.45(5.09) &
30.94(5.31) &
35.45(4.47) &
38.80(3.34) &
44.15(2.21) 
\\
DropEdge+GCN & - &
10.11(7.54) &
20.75(3.86) &
25.52(4.00) &
29.50(4.70) &
29.37(9.65) &
37.89(2.24)
\\
\hline
Meta-PN & - &
23.38(6.10) &
27.32(2.67) &
31.47(0.73) &
30.53(6.62) &
34.67(3.68) &
39.28(4.20)
\\
CGPN &- & OOM & OOM & OOM & OOM & OOM & OOM
\\
\hline

\name+GCN & 
\textbf{54.60(1.27)} &
\textbf{56.83(1.82)} &
\textbf{58.32(1.01)} &
\textbf{58.70(1.54)} &
\textbf{60.58(0.74)} &
\textbf{60.83(0.88)} &
61.11(1.70) 
\\

\name+GAT & 
54.35(1.97) &
56.29(2.65) &
56.76(1.71) &
58.04(1.81) &
60.03(1.45) &
59.95(1.59) &
\textbf{61.40(1.06)}
\\

\name+GCN w/o A & 
53.82(2.02) & 
55.89(1.30) &
57.13(1.14) & 
58.68(0.41) &
59.83(2.03) &
58.87(1.40) &
60.31(1.74) 
\\

\name+GAT w/o A &
54.13(0.71) &
55.18(2.28) &
56.50(1.75) &
57.85(1.77) &
59.70(1.72) &
57.84(2.56) &
59.89(0.91) 
\\
(improv.) & - & (+76.55\%) &
(+40.77\%) &
(+28.67\%) &
(+9.55\%) &
(+13.83\%) &
(+8.71\%)
\\
\hline
  \end{tabular}
  }
\end{table*}

\subsection{Experimental Settings}
In~\name,
we adopt Sentence-BERT~\cite{sentence-bert} (MiniLM) as LM to obtain sentence embeddings. As for LLM, we use ChatGPT~\cite{gpt3} (i.e., ``gpt-3.5-turbo'') and a basic form of prompt in Table S2 of Supplement S2
to generate samples.
We set the number of generated samples per class $M$ to be 10 across all datasets. 
The similarity threshold $\delta$ is chosen from 0.1 to 0.8 with step size 0.1.
Regarding the selection of top-$k$ node pairs in the set of edge prediction set $\mathcal{E}_P$, 
we fine-tune the values $k$ to be $1\times$, $5\times$, $10\times$, $30\times$, $50\times$, $70\times$, $100\times$, and $200\times$ of the number of generated samples $M |\mathcal{C}|$.
We use the integrated new graph as input data and conduct experiments on different shot scenarios using two backbone models, GCN and GAT,
to verify the effectiveness of our proposed paradigm.
The notation ``w/o A'' indicates the absence of an edge predictor module,
meaning that the generated labeled samples are treated as isolated points during model training.
For fair comparison, 
we report the average results on the test set with standard deviations of 5 runs for all experiments. 
We run all the experiments on a server with 32G memory and a single Tesla V100 GPU.
We provide our code and data here:
\url{https://github.com/jianxiangyu/LLM4NG}

Due to the space limitation, we move the experimental setup (Appendix~A.3),
case study of different prompt statements (Appendix B),
hyperparameter analysis (Appendix~C), and instances of generating samples using
LLMs (Appendix~D) to the appendix.

\subsection{Performance Comparison}
We perform zero- and few-shot node classification on TAGs.
We construct a $K$-shot training set, where each class includes $K$ samples, with $K$ values chosen from the set $\{0, 1, 2, 3, 4, 5, 10\}$.
For Cora and PubMed, 
we randomly sample $K \times |\mathcal{C}|$ nodes from all the nodes as the training set. 
From the remaining nodes,
we randomly select 500 nodes as the validation set and 1000 nodes as the test set.
For ogbn-arxiv, 
we follow its original partitioning, 
but we reduce the size of the training set to the $K$-shot setting as specified.
The results are reported in Table~\ref{cora_classification},
\ref{pubmed_classification}
and~\ref{arxiv_classification}.
From the tables, we have the following observations:

~(1)~In all cases,
\name~has demonstrated exceptional performance, 
particularly in low-shot scenarios. 
For example, on the ogbn-arxiv dataset,
the accuracy of \name~ achieves 56.83\% in the 1-shot setting, 
while the runner-up method achieves only 31.42\%. 
We have achieved a remarkable improvement of 76\% over the baseline model. 
Furthermore, 
our proposed paradigm \name~has significant enhancements on both GCN and GAT backbone models.
These results indicate that the supervision signals provided by the generated samples are 
particularly helpful to
the model.

~~(2)~\name~achieves impressive performance in zero-shot scenarios,
surpassing the performance of other methods even in their low-shot scenarios.
On PubMed, our model excels across various shot scenarios. Specifically, our model achieves a score of 75.36\% in the 0-shot setting, 
beating the highest score of 73.76\% obtained by other models in the 5-shot scenario.
This trend is also observed in the ogbn-arxiv dataset.
\name~provides a method that can be trained by GNNs in zero-shot scenarios.

~~(3)~Compared to `w/o A' cases,
\name~performs better.
The structural information between the generated nodes and the nodes in the raw dataset has demonstrated positive effects,
which plays a crucial role in improving model performance.
This is because the structure facilitates the propagation of supervisory signals, 
allowing labeled nodes to influence unlabeled ones, thereby enhancing the model's performance.

In conclusion,
the strong performance of~\name~highlights the effectiveness of leveraging generated samples to enhance the model's learning capabilities, especially when labeled data is extremely scarce.

\subsection{Case Study}
\subsubsection{Case Study for Comparison with LLM-based Enhanced Methods}

In this section, we compare our method with TAPE~\cite{tape}, the existing LLM-based enhanced method.
TAPE feeds titles and abstracts of papers into LLMs,
obtaining classification predictions and corresponding explanations through prompt statements.
The raw text (TA), predictions~(P), and explanations~(E) are then input into LMs to derive sentence representations.
Subsequently, these representations are input into three distinct GNN classifiers, with the final classification result determined by averaging the output probabilities of these classifiers.
For fairness, Sentence-Bert~\cite{sentence-bert} is used across all models to directly obtain sentence representations.

Table~\ref{exp:tape} displays the experimental results on ogbn-arxiv, a large benchmark dataset.
\name~demonstrates superior performance in few-shot scenarios. 
On the other hand, 
TAPE incurs substantial cost by requiring calls for all samples. 
In contrast, our method necessitates only 400 calls, achieving an improvement of over 400 times in terms of both \#calls and API cost.
It is also worth noting that
\name~and TAPE are compatible, and their combination can further improve the model performance.

\subsubsection{Case Study of Generated Samples}

In this section, we analyze the impact of 
the quantity of generated samples
and LLMs' stochasticity on the performance of the model in the Cora dataset.
The experiments conducted in this section are run only once and do not involve edge prediction between the generated nodes and the nodes in the raw dataset.

\begin{table}[t]
\centering
\caption{Comparison with LLM-based Enhanced Methods.
\underline{SUM} indicates that the cost is the sum of the two methods.}
\scalebox{0.65}{
\begin{tabular}{c|ccc|cc}
\toprule

& TA & E & TAPE & \name & TAPE+\name \\ 
\hline
0-shot & - & - & - & 54.60(1.27) & 56.80(0.13) \\
1-shot & 31.42(3.89) & 36.16(6.48) & 33.21(3.72) & 56.83(1.82) & 56.93(2.50) \\
2-shot & 40.63(3.52) & 41.36(6.48) & 41.98(6.76) & 58.32(1.01) & 58.73(1.37)\\
3-shot & 45.62(7.02) & 51.66(7.27) & 48.01(9.82) & 58.70(1.54) & 59.09(1.57) \\
4-shot & 55.30(4.08) & 53.10(5.62) & 56.62(2.82) & 60.58(0.74) & 61.24(1.39) \\
5-shot & 53.44(2.01) & 54.24(3.34) & 57.76(0.70) & 60.83(0.88) & 61.53(1.29) \\

\hline
\multirow{2}{*}{Cost} & \multicolumn{3}{c|}{\multirow{2}{*}{169,343 calls~(120.91\$)}} & 
400 calls & \multirow{2}{*}{\underline{SUM}}
\\
& & & & (0.25\$) \\
\bottomrule
\end{tabular}
\label{exp:tape}
}
\end{table}

\begin{figure}[t]
  \centering
  \begin{subfigure}[b]{0.21\textwidth}
    \centering
\includegraphics[width=\textwidth]{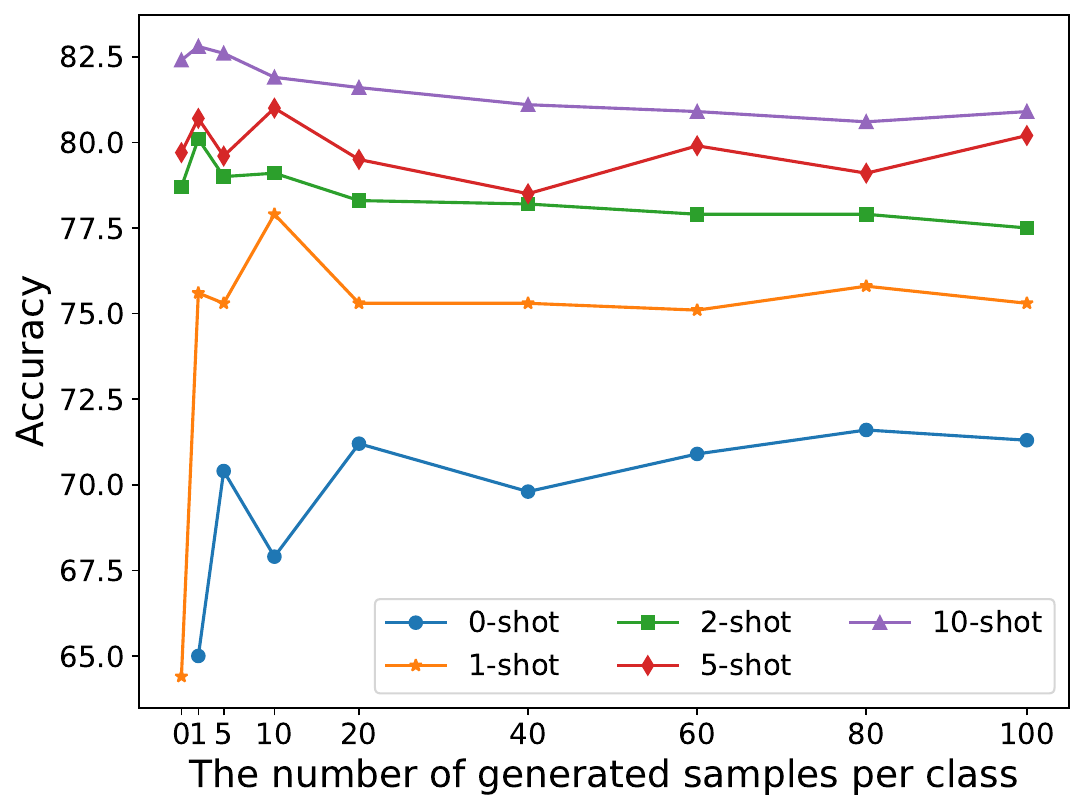}
    \caption{}
  \end{subfigure}
  \hfill
  \begin{subfigure}[b]{0.23\textwidth}
    \centering
\includegraphics[width=\textwidth]{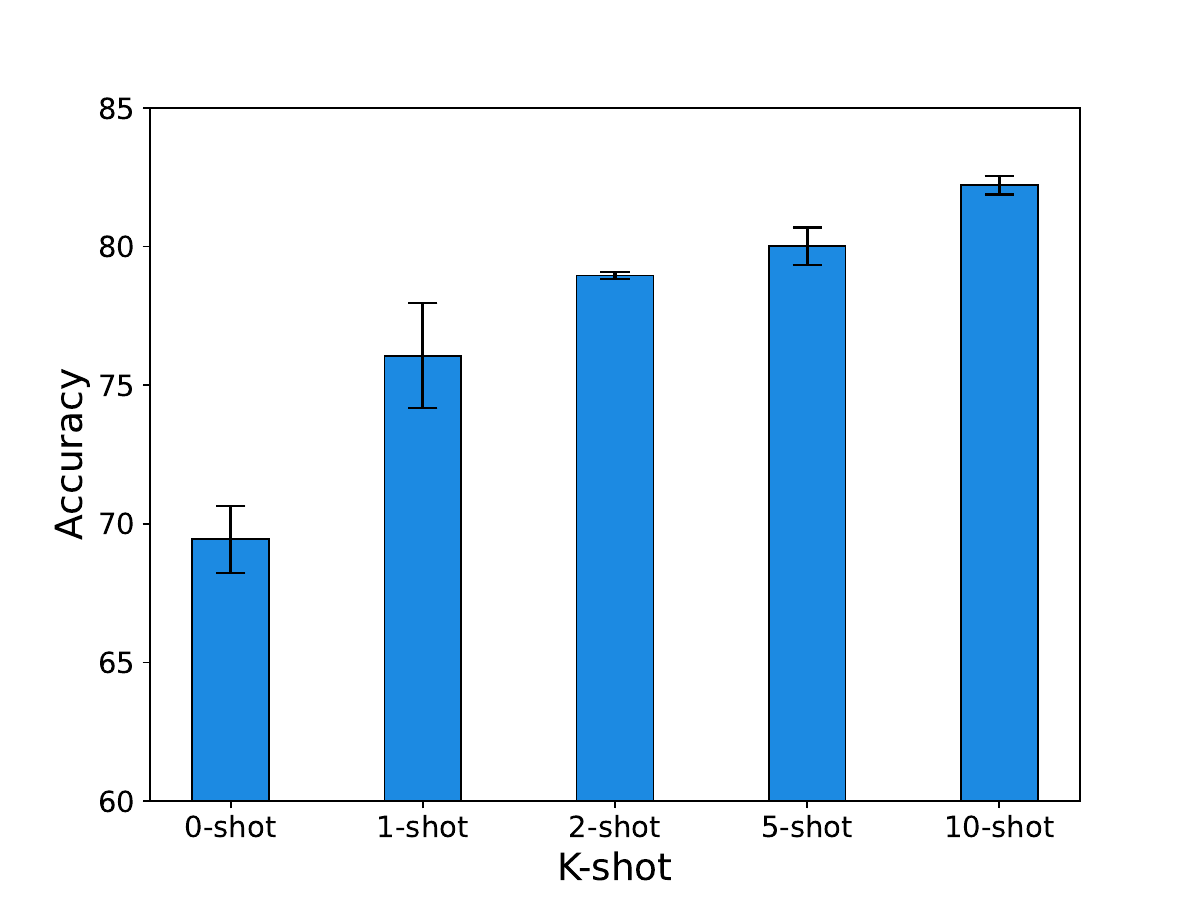}
    \caption{}
  \end{subfigure}
  \caption{Impact of (a)~the quantity of generated samples and (b)~LLMs’ stochasticity on model performance.
  }
  \label{fig:temp}
\end{figure}

~~\textbf{The quantity of generated samples.}~In this case, we use 
the same prompt statement \name-P1 to generate $100 \times 7$ samples $(M \times |\mathcal{C}|)$, and conduct experiments under different shot settings. 
Subsequently, 
we vary the number of samples generated for each class and plot the experimental results in Figure~\ref{fig:temp}(a).
We note that as the number of generated samples increases, 
the accuracy 
shows an initial increase followed by a decrease, 
and eventually stabilizes.    
This phenomenon can be explained by the increased diversity of the generated samples, 
which allows the model to obtain richer supervised signals and improve performance.
With an increasing number of generated samples,
the additional samples gradually diminish their contribution to the model's performance,
and thus the results are not significantly improved.

~~\textbf{LLMs' stochasticity.}~We also repeat the same prompt statement \name-P1 five times to generate different sample contents, 
where the number of generated samples $M$ is 10, i.e., a total of 350~($5\times10(M)\times7(|\mathcal{C}|))$ samples. 
We ensure that the generated samples are not identical to each other.
From Figure~\ref{fig:temp}(b),
we observe that although there are variations among the generated samples,
the overall quality is quite good, exhibiting strong generalization capabilities. 
We believe that the effectiveness of the generated samples by LLMs lies in their ability to capture semantic information relevant to the label text.
This semantic information is reliable and has a positive impact on classification results.

\section{Conclusion and Future Work}
In this study, we proposed a lightweight paradigm called~\name~ to address the node classification problem in few-shot scenarios. 
Specifically, we leveraged the powerful generation capabilities of LLMs to explore the semantic information of labels and generate samples. 
We then trained an edge predictor to incorporate these generated samples into the raw dataset, 
thereby providing supervision signals to the model.
The impressive experimental results strongly demonstrated the effectiveness of \name.

In our future work,
we will continue to delve into how to utilize LLMs to generate supervision signals and improve model performance.
Firstly, for other weakly supervised scenarios, 
including label imbalance and label noise, 
our goal is to leverage LLMs to supplement samples from imbalanced classes and correct mislabeled samples.
Secondly, in the case of high labeling rates, 
we will investigate how to generate more diverse samples using known labeled samples to enrich the diversity of the supervision signals and enhance the generalization ability of the model.
Overall, we will continue to explore the combination of LLMs with graph learning in various domains.

\bibliography{aaai25}

\begin{thebibliography}{44}
\providecommand{\natexlab}[1]{#1}

\bibitem[{Benamira et~al.(2019)Benamira, Devillers, Lesot, Ray, Saadi, and Malliaros}]{benamira2019semi}
Benamira, A.; Devillers, B.; Lesot, E.; Ray, A.~K.; Saadi, M.; and Malliaros, F.~D. 2019.
\newblock Semi-supervised learning and graph neural networks for fake news detection.
\newblock In \emph{Proceedings of the 2019 IEEE/ACM International Conference on Advances in Social Networks Analysis and Mining}, 568--569.

\bibitem[{Brown et~al.(2020)Brown, Mann, Ryder, Subbiah, Kaplan, Dhariwal, Neelakantan, Shyam, Sastry, Askell et~al.}]{gpt3}
Brown, T.; Mann, B.; Ryder, N.; Subbiah, M.; Kaplan, J.~D.; Dhariwal, P.; Neelakantan, A.; Shyam, P.; Sastry, G.; Askell, A.; et~al. 2020.
\newblock Language models are few-shot learners.
\newblock \emph{Advances in neural information processing systems}, 33: 1877--1901.

\bibitem[{Chen et~al.(2023)Chen, Mao, Li, Jin, Wen, Wei, Wang, Yin, Fan, Liu et~al.}]{kea}
Chen, Z.; Mao, H.; Li, H.; Jin, W.; Wen, H.; Wei, X.; Wang, S.; Yin, D.; Fan, W.; Liu, H.; et~al. 2023.
\newblock Exploring the potential of large language models (llms) in learning on graphs.
\newblock \emph{arXiv preprint arXiv:2307.03393}.

\bibitem[{Chien et~al.(2021)Chien, Chang, Hsieh, Yu, Zhang, Milenkovic, and Dhillon}]{giant}
Chien, E.; Chang, W.-C.; Hsieh, C.-J.; Yu, H.-F.; Zhang, J.; Milenkovic, O.; and Dhillon, I.~S. 2021.
\newblock Node feature extraction by self-supervised multi-scale neighborhood prediction.
\newblock \emph{arXiv preprint arXiv:2111.00064}.

\bibitem[{Devlin et~al.(2018)Devlin, Chang, Lee, and Toutanova}]{bert}
Devlin, J.; Chang, M.-W.; Lee, K.; and Toutanova, K. 2018.
\newblock Bert: Pre-training of deep bidirectional transformers for language understanding.
\newblock \emph{arXiv preprint arXiv:1810.04805}.

\bibitem[{Ding et~al.(2022)Ding, Wang, Caverlee, and Liu}]{metapn}
Ding, K.; Wang, J.; Caverlee, J.; and Liu, H. 2022.
\newblock Meta propagation networks for graph few-shot semi-supervised learning.
\newblock In \emph{Proceedings of the AAAI Conference on Artificial Intelligence}, volume~36, 6524--6531.

\bibitem[{Du et~al.(2021)Du, Qian, Liu, Ding, Qiu, Yang, and Tang}]{du2021glm}
Du, Z.; Qian, Y.; Liu, X.; Ding, M.; Qiu, J.; Yang, Z.; and Tang, J. 2021.
\newblock Glm: General language model pretraining with autoregressive blank infilling.
\newblock \emph{arXiv preprint arXiv:2103.10360}.

\bibitem[{Fan et~al.(2023)Fan, Zhao, Li, Liu, Mei, Wang, Tang, and Li}]{fan2023recommender}
Fan, W.; Zhao, Z.; Li, J.; Liu, Y.; Mei, X.; Wang, Y.; Tang, J.; and Li, Q. 2023.
\newblock Recommender systems in the era of large language models (llms).
\newblock \emph{arXiv preprint arXiv:2307.02046}.

\bibitem[{Fey and Lenssen(2019)}]{pyg}
Fey, M.; and Lenssen, J.~E. 2019.
\newblock Fast graph representation learning with PyTorch Geometric.
\newblock \emph{arXiv preprint arXiv:1903.02428}.

\bibitem[{Hafidi et~al.(2020)Hafidi, Ghogho, Ciblat, and Swami}]{hafidi2020graphcl}
Hafidi, H.; Ghogho, M.; Ciblat, P.; and Swami, A. 2020.
\newblock GraphCL: Contrastive Self-Supervised Learning of Graph Representations.
\newblock arXiv:2007.08025.

\bibitem[{Harris(1954)}]{bow}
Harris, Z.~S. 1954.
\newblock Distributional structure.
\newblock \emph{Word}, 10(2-3): 146--162.

\bibitem[{Hassani and Khasahmadi(2020)}]{mvgrl}
Hassani, K.; and Khasahmadi, A.~H. 2020.
\newblock Contrastive multi-view representation learning on graphs.
\newblock In \emph{International conference on machine learning}, 4116--4126. PMLR.

\bibitem[{He et~al.(2020)He, Liu, Gao, and Chen}]{he2020deberta}
He, P.; Liu, X.; Gao, J.; and Chen, W. 2020.
\newblock Deberta: Decoding-enhanced bert with disentangled attention.
\newblock \emph{arXiv preprint arXiv:2006.03654}.

\bibitem[{He et~al.(2023)He, Bresson, Laurent, and Hooi}]{tape}
He, X.; Bresson, X.; Laurent, T.; and Hooi, B. 2023.
\newblock Explanations as Features: LLM-Based Features for Text-Attributed Graphs.
\newblock \emph{arXiv preprint arXiv:2305.19523}.

\bibitem[{Hu et~al.(2020)Hu, Fey, Zitnik, Dong, Ren, Liu, Catasta, and Leskovec}]{arxiv}
Hu, W.; Fey, M.; Zitnik, M.; Dong, Y.; Ren, H.; Liu, B.; Catasta, M.; and Leskovec, J. 2020.
\newblock Open graph benchmark: Datasets for machine learning on graphs.
\newblock \emph{Advances in neural information processing systems}, 33: 22118--22133.

\bibitem[{Huang and Zitnik(2020)}]{huang2020graph}
Huang, K.; and Zitnik, M. 2020.
\newblock Graph meta learning via local subgraphs.
\newblock \emph{Advances in neural information processing systems}, 33: 5862--5874.

\bibitem[{Kipf and Welling(2016)}]{gcn}
Kipf, T.~N.; and Welling, M. 2016.
\newblock Semi-supervised classification with graph convolutional networks.
\newblock \emph{arXiv preprint arXiv:1609.02907}.

\bibitem[{Lan et~al.(2020)Lan, Wang, Du, Song, Tao, and Guan}]{lan2020node}
Lan, L.; Wang, P.; Du, X.; Song, K.; Tao, J.; and Guan, X. 2020.
\newblock Node classification on graphs with few-shot novel labels via meta transformed network embedding.
\newblock \emph{Advances in Neural Information Processing Systems}, 33: 16520--16531.

\bibitem[{Li et~al.(2023)Li, Liu, Fan, Wei, Liu, Tang, and Li}]{li2023empowering}
Li, J.; Liu, Y.; Fan, W.; Wei, X.-Y.; Liu, H.; Tang, J.; and Li, Q. 2023.
\newblock Empowering Molecule Discovery for Molecule-Caption Translation with Large Language Models: A ChatGPT Perspective.
\newblock \emph{arXiv preprint arXiv:2306.06615}.

\bibitem[{Liu et~al.(2022)Liu, Li, Li, Giunchiglia, Feng, and Guan}]{liu2022few}
Liu, Y.; Li, M.; Li, X.; Giunchiglia, F.; Feng, X.; and Guan, R. 2022.
\newblock Few-shot node classification on attributed networks with graph meta-learning.
\newblock In \emph{Proceedings of the 45th international ACM SIGIR conference on research and development in information retrieval}, 471--481.

\bibitem[{Liu et~al.(2021)Liu, Fang, Liu, and Hoi}]{liu2021relative}
Liu, Z.; Fang, Y.; Liu, C.; and Hoi, S.~C. 2021.
\newblock Relative and absolute location embedding for few-shot node classification on graph.
\newblock In \emph{Proceedings of the AAAI conference on artificial intelligence}, volume~35, 4267--4275.

\bibitem[{McCallum et~al.(2000)McCallum, Nigam, Rennie, and Seymore}]{cora}
McCallum, A.~K.; Nigam, K.; Rennie, J.; and Seymore, K. 2000.
\newblock Automating the construction of internet portals with machine learning.
\newblock \emph{Information Retrieval}, 3: 127--163.

\bibitem[{Miaschi and Dell{'}Orletta(2020)}]{miaschi-dellorletta-2020-contextual}
Miaschi, A.; and Dell{'}Orletta, F. 2020.
\newblock Contextual and Non-Contextual Word Embeddings: an in-depth Linguistic Investigation.
\newblock In \emph{Proceedings of the 5th Workshop on Representation Learning for NLP}, 110--119. Online: Association for Computational Linguistics.

\bibitem[{Nguyen et~al.(2020)Nguyen, Sugiyama, Nakov, and Kan}]{nguyen2020fang}
Nguyen, V.-H.; Sugiyama, K.; Nakov, P.; and Kan, M.-Y. 2020.
\newblock Fang: Leveraging social context for fake news detection using graph representation.
\newblock In \emph{Proceedings of the 29th ACM international conference on information \& knowledge management}, 1165--1174.

\bibitem[{Ni, Li, and McAuley(2019)}]{ni-etal-2019-justifying}
Ni, J.; Li, J.; and McAuley, J. 2019.
\newblock Justifying Recommendations using Distantly-Labeled Reviews and Fine-Grained Aspects.
\newblock In \emph{Proceedings of the 2019 Conference on Empirical Methods in Natural Language Processing and the 9th International Joint Conference on Natural Language Processing (EMNLP-IJCNLP)}, 188--197. Hong Kong, China: Association for Computational Linguistics.

\bibitem[{Reimers and Gurevych(2019)}]{sentence-bert}
Reimers, N.; and Gurevych, I. 2019.
\newblock Sentence-bert: Sentence embeddings using siamese bert-networks.
\newblock \emph{arXiv preprint arXiv:1908.10084}.

\bibitem[{Rong et~al.(2019)Rong, Huang, Xu, and Huang}]{rong2019dropedge}
Rong, Y.; Huang, W.; Xu, T.; and Huang, J. 2019.
\newblock Dropedge: Towards deep graph convolutional networks on node classification.
\newblock \emph{arXiv preprint arXiv:1907.10903}.

\bibitem[{Salton and Buckley(1988)}]{tfidf}
Salton, G.; and Buckley, C. 1988.
\newblock Term-weighting approaches in automatic text retrieval.
\newblock \emph{Information processing \& management}, 24(5): 513--523.

\bibitem[{Sen et~al.(2008)Sen, Namata, Bilgic, Getoor, Galligher, and Eliassi-Rad}]{pubmed}
Sen, P.; Namata, G.; Bilgic, M.; Getoor, L.; Galligher, B.; and Eliassi-Rad, T. 2008.
\newblock Collective classification in network data.
\newblock \emph{AI magazine}, 29(3): 93--93.

\bibitem[{Sun, Lin, and Zhu(2020)}]{m3s}
Sun, K.; Lin, Z.; and Zhu, Z. 2020.
\newblock Multi-stage self-supervised learning for graph convolutional networks on graphs with few labeled nodes.
\newblock In \emph{Proceedings of the AAAI conference on artificial intelligence}, volume~34, 5892--5899.

\bibitem[{Touvron et~al.(2023)Touvron, Lavril, Izacard, Martinet, Lachaux, Lacroix, Rozi{\`e}re, Goyal, Hambro, Azhar et~al.}]{touvron2023llama}
Touvron, H.; Lavril, T.; Izacard, G.; Martinet, X.; Lachaux, M.-A.; Lacroix, T.; Rozi{\`e}re, B.; Goyal, N.; Hambro, E.; Azhar, F.; et~al. 2023.
\newblock Llama: Open and efficient foundation language models.
\newblock \emph{arXiv preprint arXiv:2302.13971}.

\bibitem[{Veli{\v{c}}kovi{\'c} et~al.(2017)Veli{\v{c}}kovi{\'c}, Cucurull, Casanova, Romero, Lio, and Bengio}]{gat}
Veli{\v{c}}kovi{\'c}, P.; Cucurull, G.; Casanova, A.; Romero, A.; Lio, P.; and Bengio, Y. 2017.
\newblock Graph attention networks.
\newblock \emph{arXiv preprint arXiv:1710.10903}.

\bibitem[{Veli{\v{c}}kovi{\'c} et~al.(2018)Veli{\v{c}}kovi{\'c}, Fedus, Hamilton, Li{\`o}, Bengio, and Hjelm}]{dgi}
Veli{\v{c}}kovi{\'c}, P.; Fedus, W.; Hamilton, W.~L.; Li{\`o}, P.; Bengio, Y.; and Hjelm, R.~D. 2018.
\newblock Deep graph infomax.
\newblock \emph{arXiv preprint arXiv:1809.10341}.

\bibitem[{Wan et~al.(2021{\natexlab{a}})Wan, Pan, Yang, and Gong}]{wan2021contrastive}
Wan, S.; Pan, S.; Yang, J.; and Gong, C. 2021{\natexlab{a}}.
\newblock Contrastive and generative graph convolutional networks for graph-based semi-supervised learning.
\newblock In \emph{Proceedings of the AAAI conference on artificial intelligence}, volume~35, 10049--10057.

\bibitem[{Wan et~al.(2021{\natexlab{b}})Wan, Zhan, Liu, Yu, Pan, and Gong}]{cgpn}
Wan, S.; Zhan, Y.; Liu, L.; Yu, B.; Pan, S.; and Gong, C. 2021{\natexlab{b}}.
\newblock Contrastive graph poisson networks: Semi-supervised learning with extremely limited labels.
\newblock \emph{Advances in Neural Information Processing Systems}, 34: 6316--6327.

\bibitem[{Wang et~al.(2017)Wang, Pan, Long, Zhu, and Jiang}]{wang2017mgae}
Wang, C.; Pan, S.; Long, G.; Zhu, X.; and Jiang, J. 2017.
\newblock Mgae: Marginalized graph autoencoder for graph clustering.
\newblock In \emph{Proceedings of the 2017 ACM on Conference on Information and Knowledge Management}, 889--898.

\bibitem[{Wang et~al.(2022{\natexlab{a}})Wang, Yang, Huang, Jiao, Yang, Jiang, Majumder, and Wei}]{e5}
Wang, L.; Yang, N.; Huang, X.; Jiao, B.; Yang, L.; Jiang, D.; Majumder, R.; and Wei, F. 2022{\natexlab{a}}.
\newblock Text embeddings by weakly-supervised contrastive pre-training.
\newblock \emph{arXiv preprint arXiv:2212.03533}.

\bibitem[{Wang et~al.(2022{\natexlab{b}})Wang, Ding, Zhang, Chen, and Li}]{wang2022task}
Wang, S.; Ding, K.; Zhang, C.; Chen, C.; and Li, J. 2022{\natexlab{b}}.
\newblock Task-adaptive few-shot node classification.
\newblock In \emph{Proceedings of the 28th ACM SIGKDD Conference on Knowledge Discovery and Data Mining}, 1910--1919.

\bibitem[{Wang et~al.(2021)Wang, Wang, Liang, Cai, and Hooi}]{wang2021mixup}
Wang, Y.; Wang, W.; Liang, Y.; Cai, Y.; and Hooi, B. 2021.
\newblock Mixup for node and graph classification.
\newblock In \emph{Proceedings of the Web Conference 2021}, 3663--3674.

\bibitem[{Yang et~al.(2021)Yang, Liu, Xiao, Li, Lian, Agrawal, Singh, Sun, and Xie}]{graphformers}
Yang, J.; Liu, Z.; Xiao, S.; Li, C.; Lian, D.; Agrawal, S.; Singh, A.; Sun, G.; and Xie, X. 2021.
\newblock GraphFormers: GNN-nested transformers for representation learning on textual graph.
\newblock \emph{Advances in Neural Information Processing Systems}, 34: 28798--28810.

\bibitem[{Yao, Mao, and Luo(2019)}]{yao2019graph}
Yao, L.; Mao, C.; and Luo, Y. 2019.
\newblock Graph convolutional networks for text classification.
\newblock In \emph{Proceedings of the AAAI conference on artificial intelligence}, volume~33, 7370--7377.

\bibitem[{Zhao et~al.(2022)Zhao, Qu, Li, Yan, Liu, Li, Xie, and Tang}]{glem}
Zhao, J.; Qu, M.; Li, C.; Yan, H.; Liu, Q.; Li, R.; Xie, X.; and Tang, J. 2022.
\newblock Learning on large-scale text-attributed graphs via variational inference.
\newblock \emph{arXiv preprint arXiv:2210.14709}.

\bibitem[{Zhu et~al.(2021)Zhu, Cui, Liu, Sun, Li, Pelger, Yang, Zhang, Zhang, and Zhao}]{zhu2021textgnn}
Zhu, J.; Cui, Y.; Liu, Y.; Sun, H.; Li, X.; Pelger, M.; Yang, T.; Zhang, L.; Zhang, R.; and Zhao, H. 2021.
\newblock Textgnn: Improving text encoder via graph neural network in sponsored search.
\newblock In \emph{Proceedings of the Web Conference 2021}, 2848--2857.

\bibitem[{Zhu et~al.(2020)Zhu, Xu, Yu, Liu, Wu, and Wang}]{grace}
Zhu, Y.; Xu, Y.; Yu, F.; Liu, Q.; Wu, S.; and Wang, L. 2020.
\newblock Deep graph contrastive representation learning.
\newblock \emph{arXiv preprint arXiv:2006.04131}.

\end{thebibliography}

\appendix

\section{Appendix A Experimental Details}

\label{exp_supp}
\subsection{A.1 Datasets}
\label{dataset_description}
To evaluate the performance of~\name, we employ three
citation network datasets:
Cora~\cite{cora}, Pubmed~\cite{pubmed}, and ogbn-arxiv~\cite{arxiv}.
These datasets are citation networks that are widely used for node classification. 
In these datasets, 
each node represents a publication and the edges between nodes represent citations between the publications. 
In Cora, 
the papers belong to seven different classes within the field of computer science. These classes include Case-Based, Genetic Algorithms, Neural Networks, Probabilistic Methods, Reinforcement Learning, Rule Learning, and Theory.
The medical literature in Pubmed is divided into three classes:
Diabetes Mellitus, Experimental,
Diabetes Mellitus Type 1 and
Diabetes Mellitus Type 2.
As for ogbn-arxiv\footnote{https://ogb.stanford.edu/docs/nodeprop/},
it is a large-scale dataset of 40 classes that encompasses papers from various disciplines, 
including computer science, mathematics, social sciences, and more.
Table~\ref{dataset}~provides statistics of three datasets.

\begin{table}[ht]
\centering
\caption{Statistics of datasets}
\begin{tabular}{c|ccc}
\toprule
Dataset & Nodes & Edges & Classes \\
\hline
Cora & 2,708 & 5,429 & 7\\
Pubmed & 19,717 & 44,338 & 3\\
ogbn-arxiv & 169,343 & 1,166,243 & 40\\
\bottomrule
\end{tabular}
\label{dataset}
\end{table}

\subsection{A.2 Baselines} \label{baselines}

\noindent$\bullet$
\textbf{[Classical base models]}: GCN~\cite{gcn} and GAT~\cite{gat} are common benchmark models in graph learning for node-level representation learning based on graph convolution and attention mechanisms, respectively.

\noindent$\bullet$
\textbf{[Graph self-supervised models]}:
DGI focuses on maximizing the alignment between node representations and global summary vectors. 
Both MVGRL and GRACE use data augmentation to generate different views for contrastive learning, the former adopts graph diffusion and subgraph sampling techniques, 
while the latter employs edge dropping and node feature masking.

\noindent$\bullet$
\textbf{[Traditional graph data augmentation methods]}:
Mixup~\cite{wang2021mixup} mixes node attributes and topology to learn more discriminative features.
DropEdge~\cite{rong2019dropedge} randomly removes some edges from the input graph to improve model generalization. 

\noindent$\bullet$
\textbf{[Graph models in few-shot scenarios]}:
CGPN~\cite{cgpn} is inspired by Poisson Learning and propagates limited labels to the entire graph by contrastive learning between two networks.
Meta-PN~\cite{metapn} employs a meta-learning strategy to generate high-quality pseudo-labels, effectively enhancing scarce labeled data.

\subsection{A.3 Experiment Setups}
\label{exp_setup}
For the GCN, GAT, and \name-based methods, 
we use the same range of hyperparameters. 
For the other methods, 
we follow the hyperparameter setting in their reporsitory\footnote{https://github.com/LEAP-WS/CGPN}
\textsuperscript{,}\footnote{https://github.com/kaize0409/Meta-PN}
\textsuperscript{,}\footnote{https://github.com/PetarV-/DGI}
\textsuperscript{,}\footnote{https://github.com/CRIPAC-DIG/GRACE}
\textsuperscript{,}\footnote{https://github.com/hengruizhang98/mvgrl}
.
The search space is proivded as:
\begin{enumerate}
    \item Hidden dimension:$\{$16, 32, 64, 128, 256$\}$
    \item Dropout: $\{$0.0, 0.2, 0.5, 0.8$\}$
    \item Learning rate: $\{$1e-2, 5e-2, 5e-3, 1e-3$\}$
    \item Weight Decay: $\{$5e-4, 5e-5, 0$\}$
\end{enumerate}

\section{B Case Study of Different Prompt Statements}

We analyze the impact of 
prompt statements on the performance of the model in the Cora dataset.
The experiments conducted in this section are run only once and do not involve edge prediction between the generated nodes and the nodes in the raw dataset.
We design six different prompt statements and ensure an equal number of generated samples (i.e., 70 samples) for each case. 
These statements indicate as follows:
(1) \name-P1 is a basic form of prompt that directly generates samples related to the category.
(2) Building upon \name-P1, \name-P2 includes a description of the text generation task, introducing the roles of nodes and edges in the graph data.
(3) \name-P3 takes the samples from the training set as exemplars for LLM and generates samples based on them.
(4) \name-P4 provides all the labels for LLM to generate $|\mathcal{C}|$ samples in one go.
(5) Compared to \name-P4 based on all categories, 
\name-P5 generates $M$ samples based on a specific category.
(6) For \name-P6, 
LLM generates different samples based on previously generated samples.

In Table~\ref{prompt_statement}, we list specific examples of prompt statements. 
By comparing the experimental results in Table~\ref{prompt_statement_result} and the differences among the raw texts of the generated samples, 
we find that:

(1) Compared to the basic form of prompt~\name-P1, 
\name-P2 has an improvement in accuracy for 1-shot and 3-shot scenarios, 
indicating that the task description in the prompt enables LLM to better understand the task and generate more accurate samples related to the task. 
However, there is a decrease in accuracy in other scenarios, suggesting that the content of the task description may need further optimization.

\begin{table}[H]
  \caption{Prompt statements for LLMs to generate samples.}
  \resizebox{0.45\textwidth}{!}
  {
  \begin{tabular}{p{7.9cm}}
    \hline
    \hline
\textbf{Prompt 1:} 
\; \emph{(basic form)}
\\ 
\quad Please generate a paper belonging to category [class name], including title and abstract.
\\
\hline
\textbf{Prompt 2:} 
\; \emph{(including a description of the text generation task
)}
\\
\quad
We want to generate a few nodes from graph-structured data where the nodes represent individual research papers and the edges represent citation relationships among the papers.
\\
\quad Please generate a paper belonging to category [class name], including title and abstract.
\\
\hline
\textbf{Prompt 3:} 
\; \emph{(taking training set samples as exemplars)}
\\
\quad Here are two papers belonging to category [class name],
$[$paper1 title and abstract$]$,
$[$paper2 title and abstract$]$.
Please generate another paper belonging to category [class name], including title and abstract.
\\
\hline
\textbf{Prompt 4:} 
\; \emph{(providing labels and generate $|\mathcal{C}|$ samples at once)}
\\
\quad We have some research paper topics, namely Case Based, Genetic Algorithms, Neural Networks, Probabilistic Methods, Reinforcement Learning, Rule Learning and Theory.
\\
\quad Please generate a title and abstract for each of these topics.
\\
\hline
\textbf{Prompt 5:} 
\; \emph{(generating $M$ samples based on a specific category)}
\\
\quad Please generate 10 papers belonging to category [class name], including title and abstract.
\\
\hline
\textbf{Prompt 6:} 
\; \emph{(generating different samples based on previously generated samples)}
\\
\quad \emph{User:~}Please generate a paper belonging to category [class name], including title and abstract.
\\
\quad \emph{LLM:~} 
Title:...
Abstract:...
\\
\quad \emph{User:~}
The title of the paper you generated previously is [Generated paper1 title]. Please generate a paper belonging to category [class name], including title and abstract, which is different from the previous one you generated.
\\
\quad \emph{LLM:~} 
Title:...
Abstract:...
\\
\quad \emph{User:~}
The title of the paper you generated previously is [Generated paper1 title],[Generated paper2 title]. Please generate a paper belonging to category [class name], including title and abstract, which is different from the previous one you generated.
\\
\quad ...
\\
       \hline
    \hline
  \end{tabular}}
  \label{prompt_statement}
\end{table}

(2) In \name-P3, 
there is a significant improvement in accuracy for the 1-shot scenario, but a decrease in accuracy for subsequent shots. 
This might be because only one sample is provided as an exemplar in the 1-shot scenario, 
and the generated samples display a high degree of relevance to the raw dataset.
However, as the number of exemplars provided to LLMs increases, 
the generated samples may lack diversity.
Therefore, when providing the number of exemplars in the prompt statement, 
a balance needs to be struck between the diversity of the generated samples and their relevance to the raw dataset.

(3) \name-P4 shows a boost in each scenario, 
as providing all label information to LLM enables it to grasp the semantic information of all labels and explore the difference between them, 
thus generating more indistinguishable samples and enhancing model performance. 
However, such prompts also have limitations, as generating a large number of samples in one go may be limited by output text length, especially in scenarios with a large number of categories.

(4) \name-P5 explores the characteristics and variations of each category to generate more diverse samples, 
while \name-P6 aims to generate more distinctive samples. 
Compared to the samples generated from other prompt statements,
these two methods generate richer samples in terms of sentence structure and content. 
However, such samples may lack connection with the raw dataset or label text, resulting in no gains in accuracy.

In summary, two main factors influence the quality of generated samples by prompt statements. 
The first factor is the label information, 
as LLMs can generate more distinctive samples by uncovering the differences between labels.
The second factor is the balance between sample diversity and relevance to the raw dataset.
Samples that are related to the raw dataset can narrow the gap between the generated nodes and the original nodes, 
while more diverse samples can make the label space more distinct. 
However, there is often a trade-off between these two factors, so maintaining a balance in the prompt statement can maximize the performance of the model.

\begin{table}[hbtp]
\centering
\caption{Impact of different prompt statements on the model performance. Pi represents the i-th prompt.}
\scalebox{0.8}{
\begin{tabular}{c|cccccc}
\toprule
Prompt & 0-shot & 1-shot & 2-shot & 3-shot & 5-shot & 10-shot \\
\hline
\name-P1 & 
71.0 &
73.9 &
78.1 &
76.5 &
80.2 &
82.9
\\
\name-P2 & 
67.4 &
74.8 &
76.2 &
77.1 &
79.1 &
81.9
\\
\name-P3 & - & 
78.2 &
76.5 &
75.8 &
78.9 &
81.7
\\
\name-P4 & 
73.7 &
77.4 &
80.1 &
78.7 &
81.6 &
84.0
\\
\name-P5 & 
65.4 &
74.0 &
76.5 &
77.4 &
78.9 &
82.5
\\
\name-P6 & 
64.9 &
72.9 &
77.9 &
77.4 &
79.8 &
81.8
\\
\bottomrule
\end{tabular}
\label{prompt_statement_result}
}
\end{table}

\section{C Hyperparameter Analysis}
\label{hyper}
\begin{figure}[t]
  \centering
  \begin{subfigure}[b]{0.2\textwidth}
    \centering
\includegraphics[width=\textwidth]{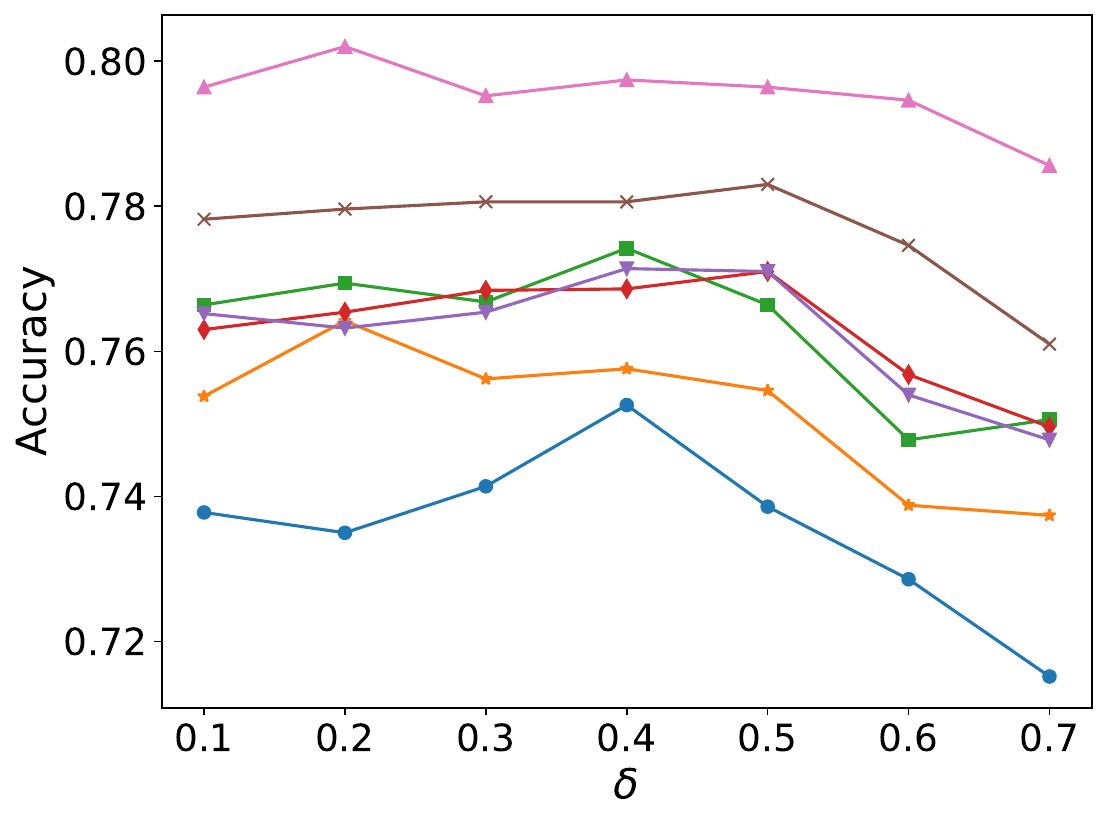}
    \caption{}
  \end{subfigure}
  \hfill
  \begin{subfigure}[b]{0.268\textwidth}
    \centering
\includegraphics[width=\textwidth]{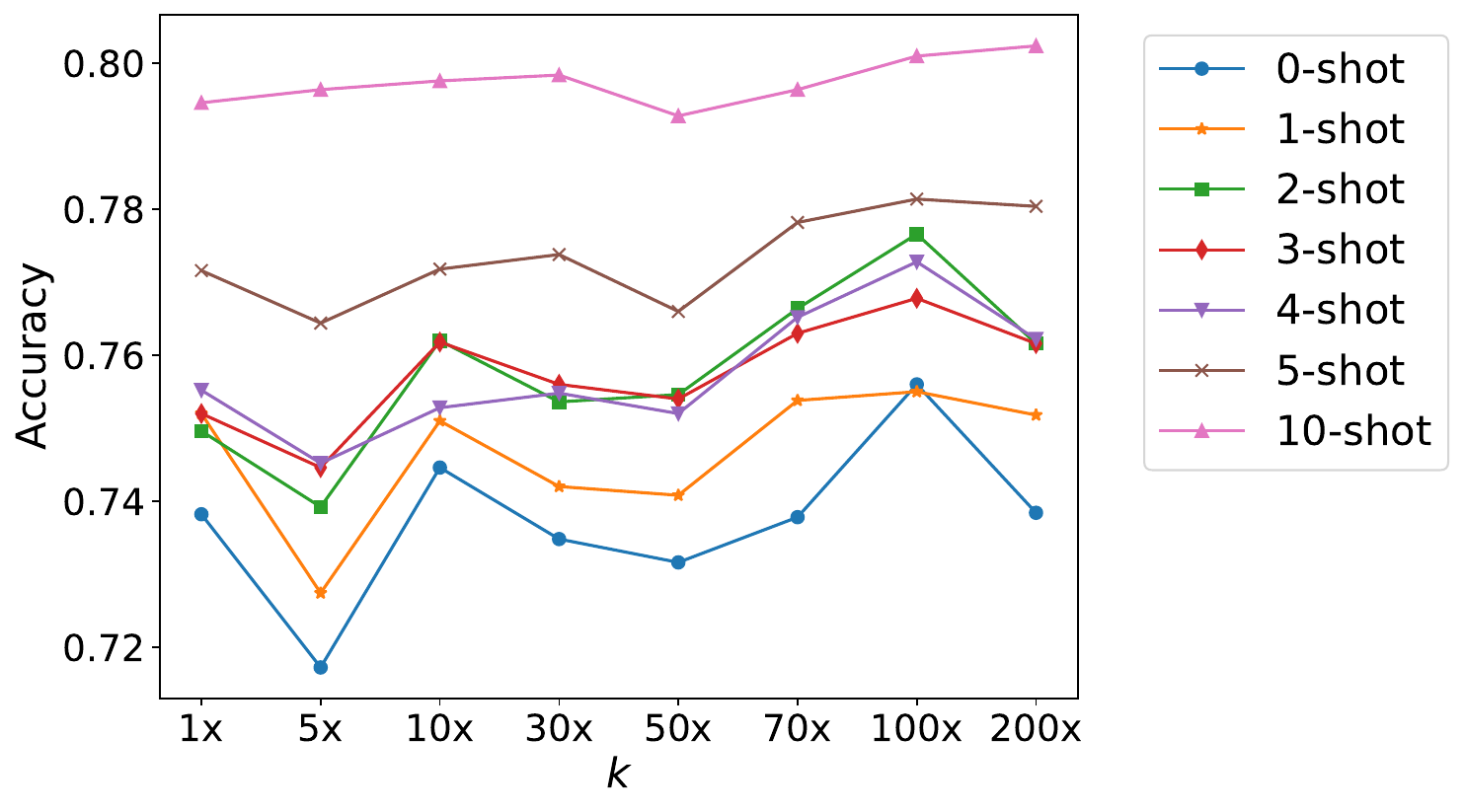}
    \caption{}
  \end{subfigure}
  \caption{Hyper-parameter sensitivity analysis on the Pubmed dataset. $\delta$ is the similarity threshold and $k$ is the number of connected edges.
  }
  \label{fig:hyper_pubmed}
\end{figure}

We further perform a sensitivity analysis on the hyperparameters of our paradigm. 
We study two main hyper-parameters in~\name:
the similarity threshold $\delta$ in Eq.4 and 
the number of edges connected between the generated nodes and the raw graph nodes $k$.
In our experiments, we vary one parameter
each time with others fixed.
From the Figure~\ref{fig:hyper_pubmed},
we see that:

(1) In the case of the threshold $\delta$,
we fix the number of edges on the Pubmed dataset by choosing 70 times the number of generated samples, i.e., 2100.
We observe that as the threshold increases, 
the accuracy initially increases and then decreases. 
This is because when the threshold is too low,
the structure includes many noisy edges,
resulting in connecting irrelevant nodes. 
These erroneous connections propagate incorrect information during message passing, 
thereby reducing the accuracy.
On the other hand, 
when the threshold is too high, 
the number of node pairs available for edge prediction in the edge set decreases. 
The newly connected nodes are already easily distinguishable samples, 
providing limited assistance for hard samples. Therefore, a high threshold restricts the benefits of structural information, leading to a decrease in accuracy.

(2) For the number of connected edges $k$, there is a consistent trend in most scenarios. 
When the number of edges is too few, 
the newly generated nodes may not have enough connections to effectively propagate the supervision signals. 
This means that the information propagation between nodes is limited, 
and the model may struggle to accurately learn the relationships and features among the nodes, thus impacting the model's performance.
Further, 
when the number of edges is too high, 
there may be an issue of over-smoothing. 
Having an excessive number of edges can lead to excessive mixing of information between node representations, 
resulting in similar node representations. 
This can confuse the model during prediction, making it difficult to accurately differentiate between different nodes and decreasing the accuracy.
In the 10-shot scenario with more supervision signals,
the discrepancy between node representations can also be recognized in the case of more edges, 
resulting in better model performance.

\begin{table*}[t]
\centering
  \caption{Samples generated with the label ``Graph''}
  \resizebox{0.95\textwidth}{!}
  {
  \begin{tabular}{c|p{15cm}}
    \toprule
\multicolumn{1}{c}{ID} & \multicolumn{1}{c}{Title} \\ 
\hline
1 & Graph-Based Recommendation Systems: An Overview and Evaluation of Algorithms.  \\
2 & A Comparative Study of Graph-Based Approaches for Community Detection in Social Networks.  \\
3 & Graph-based Analysis of Social Networks: Uncovering Community Structure and Influence Dynamics. \\
4 & Analysis of Graph Coloring Algorithms and Their Application in Network Optimization. \\
5 & Graph Analysis for Social Network Modeling and Information Diffusion.  \\
6 & Analyzing the Effectiveness of Graph-Based Algorithms for Social Network Analysis. \\
7 & The Impact of Graph Theory in Network Analysis: A Comprehensive Review. \\
8 & Graph Neural Networks: A Survey and Analysis. \\
9 & The Impact of Graph Visualization Techniques on Decision-Making: A Comparative Analysis. \\
10 & Graph Theory: Applications and Algorithms for Network Analysis. \\
\bottomrule
  \end{tabular}
  }
  \label{instances_title}
\end{table*}
\begin{table*}[t]
  \caption{An illustrative example of calling LLMs on ogbn-arxiv}
  {
  \begin{tabular}{p{16cm}}
    \hline
    \hline
\emph{User:~}
\\
\quad Please generate a paper belonging to category Artificial Intelligence, including title and abstract.
\\
\emph{LLM:~} \\
\quad Title: A Comprehensive Survey of Artificial Intelligence Techniques for Autonomous Vehicle Navigation.  
\\
\quad Abstract:
With the rapid advancements in computer vision and machine learning, artificial intelligence (AI) has emerged as a crucial component for the development of autonomous vehicles. The integration of AI techniques enables vehicles to perceive their environment, make intelligent decisions, and navigate safely through complex scenarios. This paper provides a comprehensive survey of various AI techniques employed in autonomous vehicle navigation, including perception, planning, and control. We analyze state-of-the-art approaches, discuss their strengths and limitations, and highlight ongoing research challenges in the field. Furthermore, we investigate the impact of AI advancements on improving navigation accuracy, robustness, and scalability, and outline potential future directions for the continued development of autonomous driving technology.
\\
       \hline
    \hline
  \end{tabular}}
  \label{example_ai}
\end{table*}

\section{D Instances of Generating Samples using LLMs}
\label{instances}
Table~\ref{instances_title} showcases ten instance titles generated using the basic prompt statement \name-P1 for the ``Graph'' label, 
based on the ogbn-arxiv dataset. 
We have omitted the abstract of these papers. 
From these titles, 
it is evident that LLM can capture a multitude of knowledge related to the "Graph" domain, 
such as recommendation systems, community detection, graph theory, and social networks. 
Although these papers do not exist in the real world, 
they can contain valuable knowledge about the graph domain. 
Hence, we can utilize these generated samples as supervision signals to enhance the model's performance.
In Table~\ref{example_ai}, 
we present a specific example of invoking LLM to generate samples on the ogbn-arxiv dataset. 
We can observe that the generated text is grammatically correct, logically coherent, and introduces another domain related to artificial intelligence, namely autonomous vehicle. 
Additionally, the text mentions various keywords associated with artificial intelligence, such as computer vision, machine learning, accuracy, and robustness. 
This demonstrates the ability of LLM to mine the domain knowledge of artificial intelligence.
Furthermore, 
the generated text does not require any additional processing and can be directly input into LM to obtain embeddings.

\end{document}